%% file: kdd2024.tex
\documentclass[sigconf,nonacm]{acmart}
\AtBeginDocument{%
  \providecommand\BibTeX{{%
    \normalfont B\kern-0.5em{\scshape i\kern-0.25em b}\kern-0.8em\TeX}}}

\newtheorem{hyp}{Hypothesis}
\usepackage{enumitem}
\usepackage{multirow}
\usepackage{xspace}
\setlist[itemize]{leftmargin=*}
\begin{document}

%% JH \title{Unsupervised Construction of Theme-specific Knowledge Graphs}

\title{Automated Construction of Theme-specific Knowledge Graphs}


\author{Linyi Ding}
\affiliation{%
  \institution{University of Illinois Urbana-Champaign}
  \country{IL, USA}}
\email{linyid2@illinois.edu}

\author{Sizhe Zhou}
\affiliation{%
  \institution{University of Illinois Urbana-Champaign}
  \country{IL, USA}}
\email{sizhez@illinois.edu}

\author{Jinfeng Xiao}
\affiliation{%
  \institution{University of Illinois Urbana-Champaign}
  \country{IL, USA}}
\email{jxiao13@illinois.edu}

\author{Jiawei Han}
\affiliation{%
  \institution{University of Illinois Urbana-Champaign}
  \country{IL, USA}}
\email{hanj@illinois.edu}
\renewcommand{\shortauthors}{Ding et al.}
\newcommand{\method}{TKGCon\xspace}
\newcommand{\tkg}{ThemeKG\xspace}
\newcommand{\ddd}[1]{{\color{blue} [d01: #1]}}
\newcommand{\nop}[1]{}
\newcommand\jh[1]{\textcolor{purple}{jh: #1}}
\input{1-abstract}


\begin{CCSXML}
<ccs2012>
<concept>
<concept_id>10010147.10010178.10010179.10003352</concept_id>
<concept_desc>Computing methodologies~Information extraction</concept_desc>
<concept_significance>500</concept_significance>
</concept>
<concept>
<concept_id>10002951.10002952.10002953.10002959</concept_id>
<concept_desc>Information systems~Entity relationship models</concept_desc>
<concept_significance>300</concept_significance>
</concept>
<concept>
<concept_id>10010147.10010178.10010187.10010195</concept_id>
<concept_desc>Computing methodologies~Ontology engineering</concept_desc>
<concept_significance>500</concept_significance>
</concept>
</ccs2012>
\end{CCSXML}

\ccsdesc[500]{Computing methodologies~Information extraction}
\ccsdesc[300]{Information systems~Entity relationship models}
% \ccsdesc[500]{Computing methodologies~Ontology engineering}

\keywords{Knowledge graph, theme-specific, information extraction}


\maketitle

\input{2-intro}

\input{3-related}

\input{5-method}

\input{6-experiment}

\input{7-limitation-conclution}

\clearpage
\bibliography{anthology,custom}
\bibliographystyle{acl_natbib}

% \appendix
% \input{appendix}
\end{document}

%% file: 1-abstract.tex
\begin{abstract}
 \nop{ %% Masked by JH
 Despite the comprehensive applications of knowledge graphs (KGs) in various fields such as question answering, and intelligent conversational systems, existing KGs face two major challenges: limited information granularity and deficiency in timeliness. 
 The issues considerably hinder the retrieval of in-context, fine-grained, and up-to-date knowledge from KGs, particularly in rapidly evolving contexts such as news reporting or disaster tracking. 
 }

{
Despite widespread applications of knowledge graphs (KGs) in various tasks such as question answering and intelligent conversational systems, existing KGs face two major challenges: {\em limited information granularity} and {\em deficiency in timeliness}. 
These hinder considerably the retrieval and analysis of in-context, fine-grained, and up-to-date knowledge from KGs, particularly in highly specialized themes (e.g., specialized scientific research) and rapidly evolving contexts (e.g., breaking news or disaster tracking). 
}

\nop{
 To tackle such challenges,  our work first introduces the {\em theme-specific KG} and proposes a unsupervised framework for {\bf Theme-specific KG} construction (named \method), which takes into raw theme-specific text documentations and generates a high-quality KG that includes salient entities and relations under the theme. Specifically, our framework get entity ontology of the theme from wiki and generate potential relation candidates by LLMs to enrich the ontology by relations. For documents of the theme, we first map the recognized entity pairs to the ontology and retrieve the candidate relations. Finally, we cooperate the context and ontology to extract relations.
 }

{
To tackle such challenges, we propose {\em theme-specific knowledge graph} (i.e., \emph{\tkg}), a KG constructed from a theme-specific corpus, and design an unsupervised framework for \tkg construction (named \emph{\method}).
The framework takes raw theme-specific corpus and generates a high-quality KG that includes salient entities and relations under the theme. Specifically, we start with an {\em entity ontology} of the theme from Wikipedia, based on which we then generate candidate relations by Large Language Models (LLMs) to construct a {\em relation ontology}. 
To parse the documents from the theme corpus, we first map the extracted entity pairs to the ontology and retrieve the candidate relations. Finally, we incorporate the context and ontology to consolidate the relations for entity pairs.
 }
 
\nop{
 Directly prompting GPT-4 for theme-specific KG leads to inaccurate entities (such as \textit{"two main types"}) and overly specific relations (such as \textit{"last contact with ebola patient"}). By constructing the theme-specific KG step by step, our model outperforms GPT-4 by consistently identifying accurate entities and relations. Experimental results also show that our framework excel in evaluations compared to other baseline methods.
 }

{
We observe that directly prompting GPT-4 for theme-specific KG leads to inaccurate entities (such as \textit{"two main types"} as one entity in the query result) and unclear (such as \textit{"is", "has"}) or wrong relations (such as \textit{"have due to", "to start"}). In contrast, by constructing the theme-specific KG step by step, our model outperforms GPT-4 and could consistently identify accurate entities and relations. Experimental results also show that our framework excels in evaluations compared with various KG construction baselines.
 }
\end{abstract}

%% file: 2-intro.tex
\begin{figure}[!ht]
    % \centering
    \includegraphics[width=0.95\linewidth]{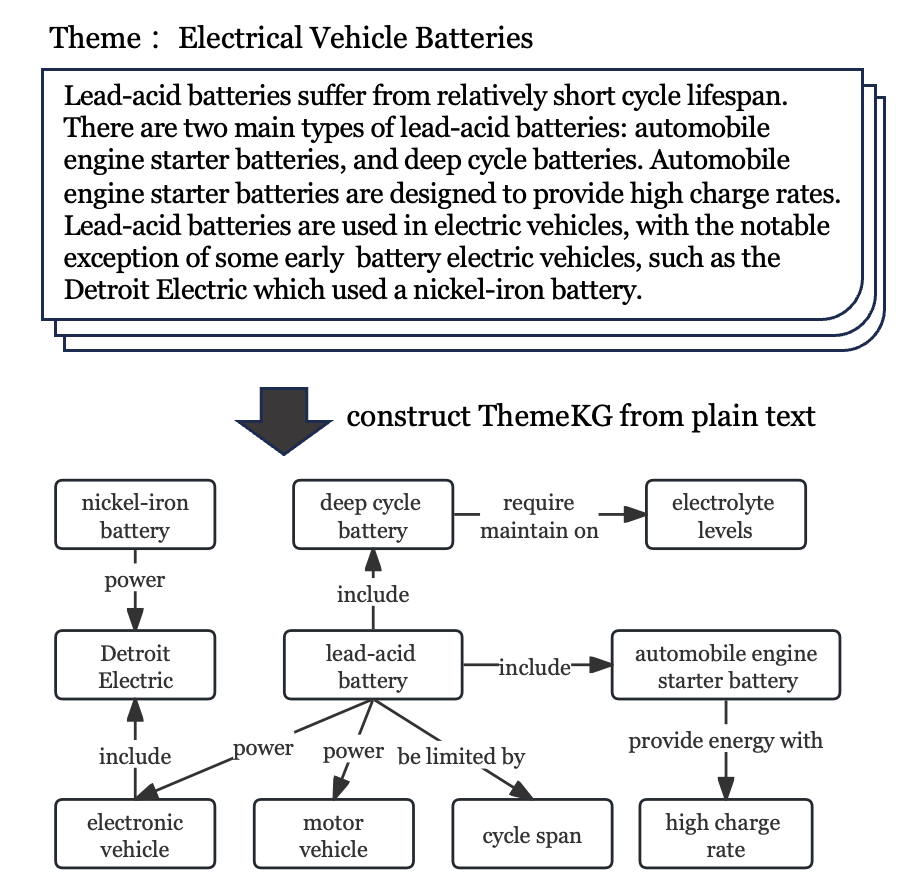}
    \caption{\method: Given a set of theme-specific documents, automatic construction of a theme-specific knowledge graph.}
    \label{fig:intro}
\end{figure}

\section{Introduction}
Knowledge graphs (KGs) \cite{ji2022survey} integrate collections of real-world entities connected by semantically meaningful relations. They usually store structured factual knowledge that allows easy access and information retrieval. 
KGs have been used in a variety of applications including recommendation systems \cite{10.1145/3209978.3210017}, question answering \cite{sun-etal-2019-pullnet}, intelligent conversational systems \cite{NEURIPS2020_6b493230}, and medical concept modeling \cite{gao2022medml}.
Existing knowledge graphs can be categorized into {\em generic open-world KGs} including Wikidata\footnote{See https://www.wikidata.org/wiki/Wikidata:Main\_Page} and {\em domain-specific KGs} including UMLS \cite{dettmers2018conve}. 

Despite the broad applications of knowledge graphs, there are two major issues attached to the existing KGs, even in the current era of large language models (LLMs). 
\nop{ %removed
One of the issues is the {\bf limited information granularity of existing KGs}.  Although the numerous sources of texts in the web may cover comprehensive and detailed information, there are specific fine-grained facts not included in the existing KGs even in domain-specific KGs. Take the theme electric vehicle battery as an instance, Wikidata contains limited and coarse-grained knowledge triples such as {\em (deep cycle battery, subclass of, rechargeable battery)}, with no regard to details of the specific theme. The fine-grained entities may also not be included, such as {\em flooded lead acid battery}.
When more detailed information such as the usage and features is queried, the existing KGs can not provide the granularity of knowledge, leading to manual retrieval in tedious web documents. LLMs also easily get hallucination on this kind of "long-tail" knowledge in a specific theme\cite{kandpal2023large}. 
}
{ % revised
The first issue is the {\bf limited information granularity of existing KGs}.
Existing KGs, including the domain-specific ones, often integrate numerous sources of texts and cover comprehensive information on a topic. They are designed for general public and do not address detailed, fine-grained information for theme-specific researchers.

\smallskip
\noindent{\bf Example 1 (EV battery)}.  Taking \emph{electric vehicle battery} as the theme, Wikidata contains limited and coarse-grained knowledge triples such as {\em $($lead-acid battery, subclass of, rechargeable battery$)$}. More finer-grained entities (e.g., {\em automobile engine starter batteries}) are not included (while it is one type of {\em lead-acid battery}).
To inquire details of such entities (e.g., usage and features), one has to rely on manual retrieval of documents. 
Also, LLMs could easily hallucinate on this kind of ``long-tail'' knowledge of a specific theme \cite{kandpal2023large}. 
}

\nop{ %removed 
Another major issue is the {\bf lack of timeliness in existing KGs}. It is practically expensive to keep those KGs in accordance with pace of the real world because the rapid emergence of themes with a number of new entities and facts. However, most of the existing KGs are not suitable for organizing up-to-date information as the construction of KGs normally requires huge efforts of human experts.
}

\smallskip
{
The second issue is the {\bf lack of timeliness in existing KGs}. 
It is hard for a KG to keep pace with the dynamics of the real world, especially for rapid changing events, since such updates often require huge efforts of human/expert annotation and guidance.
}

\nop{ % remove
When a new theme rises, such as such as {\em the international responses to Israel-Hamas conflict (2023)} , people need to retrieve news articles and organize information instead of querying existing KGs for obtaining detailed facts.
This is similar for the applications of LLMs. GPT-4\footnote{https://openai.com/research/gpt-4} suffers from the “frozen in time”  and lack of up-to-date information because of its data cuts off (September 2021).
}

\smallskip
{
\noindent{\bf Example 2 (Hamas attack on Israel)}. 
When a piece of news broke out, such as {\em Hamas attack on Israel} (Oct. 7, 2023), 
readers may like to know and digest international responses, but the existing KGs may not be updated timely. 
The knowledge of LLMs (e.g.,  
GPT-4 \cite{openai2021gpt4}) is also largely limited within the training corpus, unable to conduct smart reasoning based on the up-to-date events and information.
}
\smallskip

\nop{ % Masked
% {\em electrical vehicle battery} or {\em Hamas attack on Israel}. 
The theme-specific KG will contain 
% which may be included in other KGs. 
The challenges of constructing theme-specific KGs are the limited data of the theme and the fine-grained and up-to-date knowledge.
We observe that directly prompting GPT-4 can easily lead to factually incorrect triples 
 or improper triples. For example, the triple {\em (lead-acid battery, have, two main types)} contains inaccurate entity and unclear relation, while the triple {\em (WhiteHouse, lit up in colors of, the Israel flag)} contains overly specific and implicit relation. 
To construct KGs with salient, clear and informative entities and relations, we design an unsupervised framework for \textbf{Theme}-specific \textbf{K}nowledge \textbf{G}raph construction (named \method),  which will take in raw documents of the given theme as input and produce a quality KG on the theme with no annotations or pre-defined relations required.
Although theme-specific corpus is limited, we leverage the commonsense knowledge from Wiki and LLMs to generate a theme ontology to guide the construction. Specifically, we collect the hierarchical categories related to the theme in Wiki to construct high-level entity ontology. Then we utilize LLMs to generate potential relations candidates for pairs of entity categories as the relation ontology. 
%Guided by theme ontology, our framework  first leverages syntactic clues to extract entity spans and map the entities to the ontology. We retrieve the candidate relations based on the categories of entity pair. Finally, the relation candidates are further consolidated based on both local document contexts and the typing results. 
Guided by theme ontology, our framework types the recognized entity mentions to the theme ontology and retrieve the candidate relations in ontology for entity pairs. Finally, we cooperate the candidate relations with contextual information to further consolidate the relations.
Noted also that there are numerous works on KG constructions but few of them focus on the complete framework starting from the raw texts (which are common on the Internet and daily life) and yielding the final knowledge graph. 
}

{
To handle the above issues, this paper proposes a novel concept, \textbf{theme-specific knowledge graph (\tkg)}, that is, a knowledge graph based on a theme-specific corpus and supporting fine-grained, theme-based analysis.
This idea, though attractive, poses a major challenge: Manual construction of such graphs is too costly---we need quality, automated \tkg construction.

With the emergence of LLMs, one may wonder whether we can simply use LLM to do so. 
Unfortunately, directly prompting GPT-4 can easily lead to factually incorrect triples or improper triples. In our experiment, GPT4 generates the triple {\em (lead-acid battery, have, two main types)} containing inaccurate entity and unclear relation, and the triple {\em (WhiteHouse, lit up in colors of, the Israel flag)} containing overly specific and implicit relation. To construct KGs with salient, clear and informative entities and relations, we design an unsupervised framework for \textbf{T}heme-specific \textbf{K}nowledge \textbf{G}raph \textbf{con}struction (named \textbf{\method}),  which will take raw documents of the given theme as input and produce a quality KG on the theme with no annotations or pre-defined relations required.

Considering the limited scope of a theme-specific corpus, we leverage the commonsense knowledge from Wikipedia and LLMs to generate a theme ontology (including the entity ontology and relation ontology) to guide the construction. Specifically, we collect the hierarchical categories related to the theme in Wiki to construct high-level "entity ontology". Then we utilize LLMs to generate potential relation candidates for pairs of entity categories to form the "relation ontology". 
%Guided by theme ontology, our framework  first leverages syntactic clues to extract entity spans and map the entities to the ontology. We retrieve the candidate relations based on the categories of entity pair. Finally, the relation candidates are further consolidated based on both local document contexts and the typing results. 
Guided by the theme ontology, our framework types the recognized entity mentions to the ontology and retrieves the candidate relations from LLMs. Finally, we incorporate the candidate relations with contextual information to consolidate the relations further.
Note also that there are numerous works on KG constructions but few focus on the complete framework starting from the raw texts (which is the most typical scenario) and then yielding the final KG. An example is shown in Figure~\ref{fig:intro}.
}

In summary, this work claims the following contributions:
\leftmargini=12pt
\begin{enumerate}
    \item To our best knowledge, we are the \textbf{first to introduce the theme-specific knowledge graph construction setting} along with \textbf{two newly constructed theme-specific KG construction datasets}.
    \item We propose an \textbf{unsupervised framework that integrates the free-form generation power of LLMs and the structured guidance from a general Wiki ontology to construct a fine-grained, corpus-based, theme-coherent, and hallucination-repellent \tkg}. % theme KG}. 
    More importantly,  \textbf{our framework can be easily adapted to any other themes with the only requirement of a theme-specific raw corpus}. 
    \item Quantitative experiments and qualitative analysis show that our proposed method has achieved \textbf{advanced entity recognition and relation extraction performance} while being \textbf{coherent to themes}.
\end{enumerate}

%% file: 3-related.tex
\section{Related Work}
\subsection{Information Extraction}
\subsubsection{Relation Extraction}
Relation extraction (RE) is the task of recognizing semantic relationships between entities in the texts \cite{jain2024revisiting, gan2023synergistic, wang-etal-2023-improving-unsupervised, gupta2023comparison, li2023semi, yan2023joint, meng2023rapl}. 
Textual pattern-based approaches leverage relation-specific patterns and syntactic parsing to extract relations \cite{10.5555/3061053.3061220, batista-etal-2015-semi}.
Prompt-tuning-based approaches construct well-curated prompts to distill the hidden relational knowledge from large language models (LLMs) \cite{han2021ptr, 10.1145/3485447.3511998, chen2022relation}. 
Recent advances mainly lie within the scope of applying LLMs to assist the few-shot extraction of relations \cite{wadhwa2023revisiting, wan2023gptre} or synthesis of relational triples \cite{sun2024consistency}.
One line of research distinguishes itself by focusing on solving RE with weak or zero supervision \cite{wu2017indirect, 10.1145/3178876.3186024, 10.1007/978-3-031-43421-1_2}. 
To maximize models' relation understanding under scarce training signals, several works utilize entity typing information which allows the model to better induce the relations \cite{10.1145/3485447.3511998, 10.1007/978-3-031-43421-1_2, wei}. 

% creates learnable entity type tokens for prompt tuning. \citet{10.1007/978-3-031-43421-1_2} introduces entity typing patterns to limit the potential relation space as weak supervisions. \citet{wei} injects domain-specific hierarchical entity types for topological relation extraction which shows to bring a large improvement margin.
\subsubsection{Open Information Extraction}
Among existing RE efforts, open information extraction (OIE) \cite{10.5555/3061053.3061220,dong2021docoie,wei2023guide} is a line of work that connects closely to our task setting, as OIE extracts relation triples from text without requiring pre-defined relation sets. Traditional OIE models are based on linguistic features and statistic methods \cite{10.5555/3061053.3061220,cetto2018graphene, gashteovski2017minie}. Recently, neural OIE models \cite{kolluru2020imojie, huguet-cabot-navigli-2021-rebel-relation,DBLP:journals/corr/abs-2305-08703, fan2023efficient, pei2023abstractive} learn the extraction patterns and show promising results with supervised learning. Unsupervised OIE methods\cite{fu2023zoie} mainly utilize the power of PLM/LLMs. 
% Our work follows the unsupervised and open-vocabulary settings while focuses on the more fine-grained and theme-specific triple extraction.
Our work further extends unsupervised OIE to a new setting that requires the extracted entities and relations to be relevant to a specific theme, which is especially challenging when no labeled data is available.

\nop{
Our work focus on a open v

Within the scope of this paper, we designate these works as {\em Entity-augmented Relation Extraction} which refers to approaches that extract relevant information of the entities to augment or assist the task of relation extraction. 

Compared with other approaches, our work is closely related to the entity-augmented relation extraction framework while we fully exploit the large-scale entity ontologies from existing knowledge bases (i.e., Wikipedia), instead of using a handful of entity types. 
}
%Our method is able to extract open, comprehensive and well-structured entity-augmented information. 

\vspace{-1ex}
\subsection{KG Construction}

\subsubsection{LLM-Enhanced KG Construction}
Large language models \cite{brown2020language, ouyang2022training} have been shown as versatile tools for various natural language processing applications. 
Recent studies \cite{zhu2023llms, chen2023knowledge, trajanoska2023enhancing, Fang2024MolTCTM, wei2024kicgpt, li2024contextualization} have explored leveraging LLMs for KG construction. \citet{kggpt} utilizes prompting methods to convert plain text to a KG. \citet{zhu2023llms} proposes AutoKG which adopts a multi-agent-based approach employing LLMs for KG construction and reasoning while some other studies directly query LLM to reason the relations between certain entities \cite{chen2023knowledge}. 
\citet{ye2023beyond} introduces CooperKGC which establishes a collaborative processing network based on LLM to integrate entities, relations and extracted events for KG construction. \citet{jiang-etal-2023-text} generates query prompts to retrieve external information and probe knowledge from PLM for KG completion. 

\vspace{-1ex}
\subsubsection{Domain-Specific KG Construction} In recent years, 
constructing domain-specific KGs has gained significant attention  \cite{urchade2024autoregressive}.
Traditional methods involve information extraction from diverse sources \cite{yao2022schema}, while recent advancements leverage LLMs for entity and relation recognition.
Specialized KGs tailored to specific domains, such as geospatial or finance, have been constructed using domain-specific ontologies and data sources, with emphasis on noise filtering in extraction, ontology alignment and schema integration.
\cite{gan2023making} introduces a KG construction method tailored for online marketing which replaces the relation generation of LLM with a relation filtering process empowered by domain-specific knowledge. They also leverage progressive prompting augmentation for entity expansion. 
\cite{karim2023large} focuses on the biomedical domain and constructs a domain ontology for validating gene-disease relations.
They further fine-tune the biomedical KG using LLMs based on more recent articles and new knowledge bases.

Most works for domain-specific KG construction are costly in human annotations and ontology construction but are difficult to generalize to other domains. Our work focuses on automatically constructing more fine-grained, theme-specific KGs , and the framework could be easily applicable to other themes.

%our work focuses on theme-specialized, up-to-date corpus, usually containing open relations and new fine-grained entities, which is more challenging but more realistic in applications. However, most previous KG construction methods are limited to certain pre-defined sets of entity ontologies or relations. 

% \subsection{Evaluation of Constructed KG}

%% file: 5-method.tex
% \vspace{-1ex}

\begin{figure*}[!ht]
    \centering
    \includegraphics[width=0.94\linewidth]{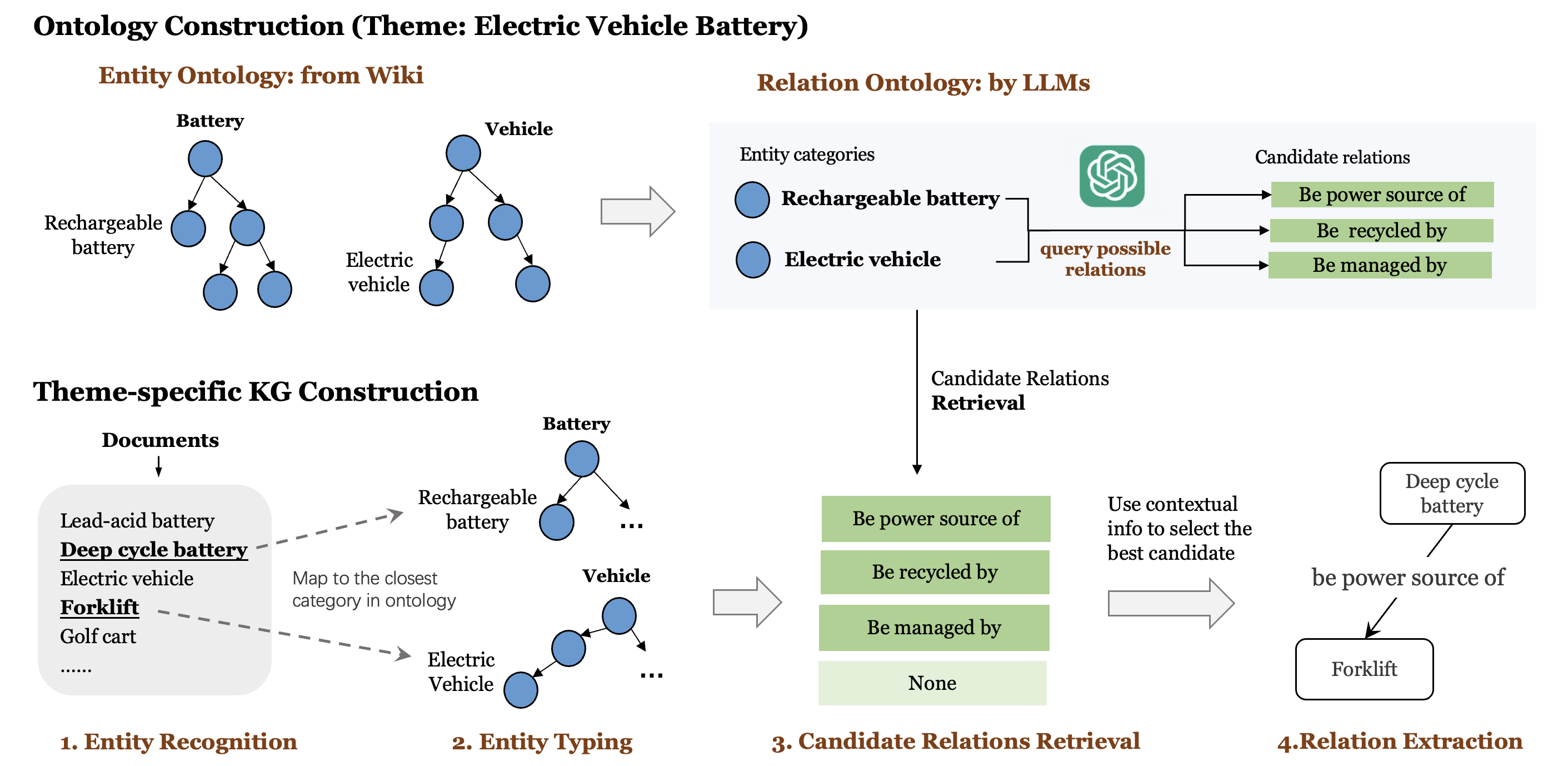}
\caption{The overall framework of \method consists of (i) ontology construction and 
(ii) \tkg construction. 
{\normalfont
For (i), it leverages the large general Wikipedia and GPT-4's reasoning ability to obtain high-quality entity ontology and relation ontology for the given theme. % Specifically, it leverages the human-curated entity ontology from Wiki and reasons the potential relations between two categories on the ontology, which are used to guide the KG construction.  
For (ii), we first process the theme documents, with phrase mining by SpaCy, entity typing by ZOE to retrieve candidate relations. Finally, the candidate relations generated by LLMs are further filtered with contextual information to consolidate final relations.}
}
    \label{fig:overview}
\end{figure*}

\section{Method}
\subsection{Task Formulation}
\subsubsection{Theme-specific Knowledge Graph (\tkg)}
Compared with existing knowledge graphs including general KGs and domain-specific KGs (focusing on a broader domain, such as biomedical sciences, sports, and finance), the theme-specific knowledge graphs (i) focus on narrower themes (such as EV-battery and Hamas-attack-on-Israel) that may need in-depth exploration for special purposes; (ii) contain more detailed knowledge including finer-grained entities and triples. For example, \emph{(deep cycle batteries, be power source of, forklifts)} can be a fine-grained triple in \tkg, while a relevant triple in general KGs could be {\em (EV batteries, be used in, electric vehicles)}. These finer-grained triples might be seen as long-tail knowledge and discarded in previous KGs.

\subsubsection{Theme-specific Knowledge Graph Construction}
Given a specific theme and a set of documents $\mathcal{D}$ with each document $d\in\mathcal{D}$ describing relevant content about the theme, our task aims to extract the theme-related knowledge triples from $\mathcal{D}$ in the format of $\left(e_i,r_k,e_j \right) $, where $e_i,e_j\in\mathcal{E}$ are salient theme-specific entities recognized from $\mathcal{D}$, and $r_k\in\mathcal{R}$ are open-vocabulary phrases that describes the relation of two entities. Here, $\mathcal{E}$ denotes the entity set, and $\mathcal{R}$ denotes the relation set under the theme.

\medskip
\noindent {\bf Example 3 (``EV battery'' \tkg construction)}. Given the theme ``{\em EV-battery}'' and the following text: ``{\em Deep cycle batteries are used to provide continuous electricity to run electric vehicles like forklifts}'', the output of theme-specific knowledge graph construction may include the following possible knowledge triples:  \textit{$($deep cycle batteries, provide, continuous electricity$)$, $($deep cycle batteries, be power source of, electric vehicles$)$, $($deep cycle batteries, be power source of, forklifts$)$, $($electric vehicles, include, forklifts$)$}.
% For a more concrete example, we show one document from the experiments in Figure~\ref{fig:intro}.
Figure~\ref{fig:intro} also shows a set of triples (presented as a knowledge graph) extracted from theme corpus by our \method approach.

\subsection{Overall Framework}
To handle the task of theme-specific knowledge graph construction, our proposed \method includes two main parts: {\em Theme Ontology Construction} and {\em Theme KG Construction}. The overall framework is shown in Figure~\ref{fig:overview}. 
\begin{itemize}
\item \textbf{Theme ontology construction}: Given a specific theme, we construct an ontology including both entities and relations. Specifically, we first collect a hierarchy of entity categories relevant to the theme from Wikipedia\footnote{https://www.wikipedia.org/} as the {\em entity ontology} (Sec.~\ref{sec:ontology-from-wiki}). Then, for each pair of entity categories in the entity ontology, we generate potential candidate relations by querying LLMs to construct the relation ontology  (Sec.~\ref{sec:potential-relations-LLM}). 
    
\item \textbf{Theme KG construction}: Given the theme-specific documents, we first mine the possible entity mentions and map them to the closest category on the entity ontology (Sec.~\ref{sec:entity-recognition-typing}). Then, for each pair of entity mentions, we retrieve the candidate relations (from the relation ontology) based on the entity categories that they map to. Finally, we incorporate the contextual information to decide the relation between these two entity mentions (Sec.~\ref{sec:relation-extraction}).
\end{itemize}

\subsection{Theme Ontology Construction}
Directly constructing \tkg from plain text can be challenging since the theme-specific corpus may lack essential commonsense knowledge for open-vocabulary information extraction. Without any guidance, the extracted triples can contain ambiguous entities (such as "two main types") or unclear relations (such as "are") or be inconsistent with commonsense knowledge.
To circumvent these issues, this paper leverages the advantages of large-scale general Wikipedia and LLM's reasoning ability to construct a {\em theme ontology} before extracting the KG triples. The theme ontology includes an {\em entity ontology} constructed from Wikipedia and a {\em relation ontology} with high-quality relation candidates generated by LLMs.

\subsubsection{Entity Ontology Construction from Wikipedia} \label{sec:ontology-from-wiki}
Wikipedia is a large online database containing rich commonsense knowledge. Although it lacks fine-grained details for theme-specific tasks, the categorization function in Wikipedia\footnote{https://en.wikipedia.org/wiki/Wikipedia:Contents/Categories} can {\em provide a high-quality, human-curated hierarchy of high-level concepts/categories about the theme in a top-down manner.}  For any entity (such as {\em flooded lead-acid battery}) that is not included in Wikipedia, we can still find its category path {\em (battery $\rightarrow$ rechargeable battery $\rightarrow$ lead-acid battery)} from the hierarchy and use it as external knowledge of the theme. This hierarchy is referred as the ``entity ontology'' where the nodes are high-level concepts/categories from Wiki.

We show a concrete example for entity ontology construction. Given a specific theme such as {\em electrical vehicle battery}, we choose the relevant concepts in the categorization of Wikipedia as initial root nodes: {\em Battery (Electricity)} and  {\em Vehicles}. Then we collect their sub-trees on the pages of Wiki Category as the basic entity ontology for the theme, including the nodes like {\em battery inventors, rechargeable batteries, battery chargers}. Each child node in the ontology denotes a sub-category. Note that the original ontology from Wiki may contain irrelevant connections, and thus we filter out wrong connections by the similarity of parent and child nodes calculated by a pre-trained transformer, all-MiniLM-L6-v2 \footnote{https://huggingface.co/sentence-transformers/all-MiniLM-L6-v2} \cite{vaswani2017attention} (which will be used again later).
In the step of \tkg construction, the entity ontology can be further expanded 
% by trees of categories as 
as entities from new categories are discovered in the documents.

\subsubsection{Relation Ontology Construction by LLMs} \label{sec:potential-relations-LLM}

Directly generating relations for an entity pair under a theme-specific context can result in undesired levels of variations or even mistakes. For example, the generated relations can be overly general (e.g., \textit{"is"}),  overly specific (e.g., \textit{"be the oldest and cheapest battery of"}), overly implicit (e.g., \textit{"lit up in colors of"}), or wrong (e.g., {\em "start"} in the triple \textit{(charging rate, start, engine)}). We observe that relations generated from entity categories are usually more robust than those generated from theme-specific entities. To construct a KG with salient, clear, and informative relations, we make the following hypothesis:  

% Open vocabulary relation extraction is mostly treated as a generative task, which may vary greatly in expression. The generated relations can be overly general (e.g., \textit{"is"}) or overly specific (e.g., \textit{"be the oldest and cheapest battery of"}) or implicit (e.g., \textit{"lit up in colors of"}). The irregularity in expression makes the relations difficult for clustering and organizing. Additionally, limited theme documents may lead to unreasonable triples, e.g., \textit{(charging rate, start, engine)}, due to a lack of commonsense knowledge, especially for parsing long and complex sentences. Therefore, we prepare the potential relation sets for possible entity pairs in advance to prevent generating improper or incorrect relations. 

%We make the following hypothesis:

\begin{hyp} Give entity categories $E_1$ and $E_2$ from Wiki ontology, the set of possible relations from entity $e_1 \in E_1$ to entity $e_2 \in E_2$, is a subset of possible relations from category $E_1$ to category $E_2$.
\end{hyp}

For example, we know that the possible relations from category \textit{rechargeable battery} to category \textit{electric vehicles} could be \{{\em "be power source of", "be recycled from", "be managed by"}\}. Then, we can reason that the relation from $e_1\in E_1$ (e.g., {\em deep cycle battery}, which is a type of {\em rechargeable battery}) to $e_2\in E_2$ (e.g., {\em forklift}, which is a type of {\em electric vehicles}) should be within the set.

Thus, we construct the ``relation ontology'' in the following approach. For a given pair of entity categories $(E_1,E_2)$ in the entity ontology collected from Wiki, we follow the hypothesis and query LLM to reason the potential candidate relations\footnote{We query in both directions: (entity 1, relation, entity 2) and (entity 2, relation, entity1).}. The final relation ontology is a map from category pairs to candidate relation sets: $(E_1,E_2)\rightarrow\mathcal{R}$.
The prompt we used is as follows:
\nop{
\vspace*{.5ex}
{\em Given the theme [Theme], what are the possible relations from [Category 1] to [Category 2]? List Answers in the format: ([Category 1], \_\_\_ , [Caterogy 2])}
\vspace*{.5ex}
 }
\begin{quote}
    {\em Given the theme [Theme], what are the possible relations from [Category 1] to [Category 2]? List Answers in the format: ([Category 1], \_\_\_ , [Caterogy 2])}
\end{quote}
Here, the commonsense knowledge in LLM can compensate for the deficiency of limited theme-specific data. For an up-to-date specific theme (such as {\em Hamas-attack-on-Israel}), the approach can reason the potential relations as long as the entity's affiliated category is known to the pre-trained LLM. Note that, we also limit the depth of Wiki entity ontology to prevent overly fine-grained entity categories, which may cause hallucination during querying LLM.

\subsection{Theme KG Construction}
We observe that end-to-end methods for KG construction tend to miss triples especially for document-level extraction. Our framework can effectively avoid the issue by following a pipeline that first recognizes the entities and then extracts their relations.

\subsubsection{Entity Recognition and Typing} \label{sec:entity-recognition-typing}  Given a set of documents about the given theme, our \method first extracts the possible entity mentions, which will be mapped to the categories on the entity ontology later. Due to the limited data and unseen fine-grained entities, we find that most existing phrase/entity extraction methods such as AutoPhrase \cite{shang2018automated} have low recall rates. To ensure a high recall rate, our approach first gets noun chunks by SpaCy \cite{spacy2} as candidate entity mentions. We then filter out the noisy mentions (e.g. \textit{"itself", "the features"}) or noisy modifiers (e.g. \textit{"most common"} in the phrase \textit{"most common vehicle batteries"}) by a set of rules.
\begin{enumerate}
    \item The valid mentions should contain at least one noun and no pronouns/stopwords according to the POS. 
    \item The phrases that have a high frequency in large corpus\footnote{https://github.com/IlyaSemenov/wikipedia-word-frequency} and low theme coherence (similar calculation could refer to Eqn.~\eqref{eq:theme-coherence}) are treated as noisy mentions (e.g. \textit{"cases"}).
    \item The non-nouns that have high frequency and low co-occurrence with other words should be the noisy modifiers (e.g. \textit{"actual"}).
\end{enumerate}

With the entity mentions from theme documents, we then map them to the closest Wiki category. We consider two cases. 
\begin{itemize}
    \item {\bf Case 1:} For a mention that directly matches a Wikipedia page title, we consider the page title as an entity and the categories at the bottom of the page as the associated categories. We then calculate two scores for determining which category from the set is the best match: (i) \textit{self coherence} between entity $e$ and category $E$; (ii) \textit{theme coherence} between theme $t$ and category $E$.
    \begin{align}
    C_{self} &= \cos \left(Rep(E),Rep(e)\right), \\
    C_{theme} &= \cos \left(Rep(E),Rep(t)\right). \label{eq:theme-coherence}
    \end{align}
    The representations $Rep(\cdot)$ are provided by the pre-trained transformer \cite{vaswani2017attention}. After obtaining $C_{theme}$, we first use a threshold to filter out categories that are irrelevant to the theme. If there is no category left after the filtering, we consider the entity mention is also irrelevant to the theme; Otherwise, we sort the multiplication value $C_{self}\times C_{theme}$ for each remaining category in the set and find the best match that corresponding to the max value.
    \vspace{1mm}
    \item {\bf Case 2:} If the mention does not have a direct match in Wikipedia, we do the following. For the entity mention $e$ and context $c$, we adapt ideas from ZOE \cite{DBLP:conf/naacl/PetersNIGCLZ18} to type the entity based on the entity ontology. Specifically, for every category $E$ in the ontology, we use the pre-trained ELMo \cite{peters-etal-2018-deep} to calculate the context consistency score $P(E\mid e, c)$ as ZOE does. If the score reaches a lower threshold, meaning that we could not find a proper match in the existing entity ontology. In this case, we then apply explicit semantic analysis \cite{gabrilovich2007computing} to fast retrieve relevant Wikipedia categories as the candidate set. Similarly, we adopt the self coherence and theme coherence scores to determine the best-matched category or consider the entity irrelevant to the theme if none of the categories in the candidate set has high $C_{theme}$.
\end{itemize}
  
\subsubsection{Relation Retrieval and Extraction} \label{sec:relation-extraction}
Given two entity mentions, we introduce how to extract their relations by incorporating the contextual information. For a pair of entity mentions $(e_1,e_2)$, assuming their typed (i.e., best matched) categories are $(E_1,E_2)$, we first retrieve candidate relations from theme relation ontology. To improve the accuracy, diversity, and coverage, we also retrieve the relations based on the parent nodes of $E_1$ and $E_2$ in the entity ontology and merge these candidate relations together.

Then, we select the best suitable relation based on the context (neighboring sentences that include the entities) by LLMs since relation extraction requires semantic understanding and reasoning abilities. Before doing so, we add the \textit{none} relation into the candidate relations in case the pair of entity mentions cannot find suitable relation in candidate set according to the context. The prompt we use is as follows:
\begin{quote}
{\em Please choose the most proper relation in the candidate set for [entity1] to [entity2] according to the context. If all the relations in the candidate set are not suitable, please choose none. The output format should be (entity1, relation, entity2). Context: [context]. Relation candidates: [relation1, relation2, \ldots, none]}
\end{quote}
\nop{ 
\vspace*{.5ex}
{\em Please choose the proper relation in candidate set of relations for [entity1] to [entity2] according to the context. If there is no relation between two entities in the context, please choose none. The output format should be (entity1, relation, entity2). Context: [context]. Relation candidates: [relation1, relation2, relation3, \ldots, none].
} 
\vspace*{.5ex}
}
If the output relation is \textit{none}, then either (i) two entities do not have factual relations from the document; or (ii) during relation ontology construction, the LLMs have missed the relation between these two entities (the cases are rare).
For both cases, without further guidance, we query LLMs again to directly extract the entity relation based on the context or output a {\em none}. This prompt is shown in Sec~\ref{sec:ablation}. If LLM gives {\em none} as the answer, we consider it as case (i); Otherwise, we generate a triple {\em (entity1, relation, entity2)} and add this new relation to enrich the theme relation ontology. 

%Our relation retrieval and extraction method can effectively avoid different expressions of the same relation. 
Finally, we organize these valid triples to get a theme-specific knowledge graph (\tkg).

%% file: 6-experiment.tex
\section{Experiment}
Our framework aims to construct theme-specific KGs from scratch with limited documents related to the given theme. The task is closely related to document-level open information extraction. This section evaluates the performance of our \method against existing baseline models on constructing two theme-specific KGs.

\begin{table} \small
    \centering
    \caption{Statistics of Datasets}
    \vspace{-3mm}
    \begin{tabular}{c|c c c c}
        \toprule
        {\bf Datasets} &  {\bf Documents} & {\bf Entities} & {\bf Relations} & {\bf Triples} \\
        \midrule
        EVB  & 20 & 130 & 64 & 330 \\
        HAI & 20 & 142 & 77 & 425 \\
        \bottomrule
    \end{tabular}
    \label{tab:datasets}
     \vspace{-1ex}
\end{table}

\subsection{Dataset Collection}
To verify the validity of our framework, we choose two themes and collect corpus for theme-specific KG construction: \textit{EV battery (EVB)} and \textit{Hamas-attack-on-Israel (2023) (HAI)}. Specifically, for each theme, we collected the documents from online sources with four criteria: (i) the relevance to the given theme, (ii) the reliability of the source contents, (iii) the granularity of description details, and (iv) the diversity of information about the theme.
 
\textit{Electrical vehicle batteries (EVB)} is a traditional theme in the industry while the specific terms or other details of this theme are hardly included in existing KGs. \textit{Hamas-attack-on-Israel (2023) } is an up-to-date theme in the press with a number of new entities (e.g. \textit{"Nova music festival"}) not included in existing knowledge bases. For each theme, we mainly collect 20 high-quality and representative documents related to the theme from various online databases (e.g. New York Times, CRS Reports\footnote{https://crsreports.congress.gov/}).
Then, we annotate the entities and triples at the document level from two corpora to generate the gold theme-specific knowledge graphs for evaluation. The statistics of two datasets EVB and HAI can be found in Table~\ref{tab:datasets}.

% \begin{table*}
%     \centering
%     \begin{tabular}{c|c|c c c|c}
%         \hline
%         Dataset & Method &  Triple Recall & Triple Precision & Triple F1 & Theme Coherence  \\
%         \hline
%         \multirow{5}{*}{EVB} & OpenIE  & 0.13& 0.24 & 0.17& 0.46 \\
%         ~ & REBEL & 0.11& 0.80& 0.19& 0.80 \\
%          ~ & IMoJIE  & 0.26 & 0.45 & 0.33 & 0.78   \\
%         ~ & KG-GPT & 0.67& 0.64&0.65& 0.95 \\
%         ~ & GPT-4  & 0.64& 0.65&0.64& 0.97  \\
%          ~ & TKGCon(w/o ontology)  & 0.67& 0.57& 0.62 & 0.92  \\
%          ~ & TKGCon  & \textbf{0.78}& 0.73&\textbf{0.75}& \textbf{0.97} \\
%         \hline
%         \multirow{5}{*}{HAI} & OpenIE  & 0.17&0.22 & 0.19& 0.35 \\
%         ~ & REBEL & 0.15& 0.75& 0.25& 0.75  \\
%          ~ & IMoJIE  & 0.25 & 0.31 & 0.28 & 0.83   \\
%         ~ & KG-GPT & 0.72& 0.69&0.70& 0.91  \\
%         ~ & GPT-4  & 0.70& 0.72&0.71& 0.93 \\
%         ~ & TKGCon(w/o ontology)  & 0.75& 0.62& 0.68 & 0.88  \\
%          ~ & TKGCon & \textbf{0.81}& \textbf{0.75}&\textbf{0.78}& 0.92  \\
%         \hline
%     \end{tabular}
%     \caption{Comparison with baselines on KG construction}
%     \label{tab:baseline1}
%     \vspace*{-2ex}
% \end{table*}

\begin{table*}
    \caption{Comparison with baselines on KG construction.}
    \vspace{-2mm}
    \centering
    \resizebox{0.80\textwidth}{!}{\begin{tabular}{c|c|c c c|ccc|c}
        \toprule
        \multirow{2}{*}{\bf Dataset} & \multirow{2}{*}{\bf Method} & \multicolumn{3}{c|}{\bf Entity Metric} & \multicolumn{3}{c|}{\bf Triple Metric} & {\bf Theme Metric} \\
        \cmidrule{3-9}
        & & {\bf Recall} & {\bf Precision} & {\bf F1-score} & {\bf Recall} & {\bf Precision} & {\bf F1-score} & {\bf Coherence} \\
        \midrule
        \multirow{5}{*}{EVB} & OpenIE \cite{10.5555/3061053.3061220}  & 0.62 & 0.36 & 0.46 & 0.13& 0.24 & 0.17& 0.46 \\
        ~ & REBEL \cite{huguet-cabot-navigli-2021-rebel-relation} & 0.22 & {\bf 0.80} &0.35 & 0.11& {\bf 0.80} & 0.19& 0.80 \\
         ~ & IMoJIE \cite{kolluru2020imojie}  & 0.44 & 0.49 & 0.46 & 0.26 & 0.45 & 0.33 & 0.78 \\
        ~ & KG-GPT \cite{kggpt} & 0.72 & 0.69 &0.70 & 0.67& 0.64&0.65& 0.95  \\
        ~ & GPT-4 \cite{openai2021gpt4}  & 0.68 & 0.71 &0.69 & 0.64& 0.65&0.64& \textbf{0.97} \\
         ~ & \method (w/o ontology)  & /&/ &/ & 0.67& 0.57& 0.62 & 0.92  \\
         ~ & \method  &\textbf{0.92} & {\bf 0.80} & \textbf{0.86} & \textbf{0.78}& 0.73&\textbf{0.75}& \textbf{0.97} \\
        \midrule
        \multirow{5}{*}{HAI} & OpenIE \cite{10.5555/3061053.3061220}  & 0.52 & 0.28 &0.36 & 0.17&0.22 & 0.19& 0.35 \\
        ~ & REBEL \cite{huguet-cabot-navigli-2021-rebel-relation} & 0.16 & 0.87 &0.27 & 0.15& 0.75& 0.25& 0.75   \\
         ~ & IMoJIE \cite{kolluru2020imojie}  & 0.33& 0.39 & 0.36  & 0.25 & 0.31 & 0.28 & 0.83  \\
        ~ & KG-GPT \cite{kggpt} & 0.84 & 0.79 &0.81 & 0.72& 0.69&0.70& 0.91 \\
        ~ & GPT-4 \cite{openai2021gpt4} & 0.82 & 0.80 &0.83 & 0.70& 0.72&0.71& \textbf{0.93}\\
        ~ & \method (w/o ontology)  & /&/&/ & 0.75& 0.62& 0.68 & 0.88 \\
         ~ & \method & \textbf{0.90} & \textbf{0.88} &\textbf{0.89} & \textbf{0.81}& \textbf{0.75}&\textbf{0.78}& 0.92  \\
        \bottomrule
    \end{tabular}}
    \label{tab:baseline1}
    \vspace*{-2ex}
\end{table*}

%We have provided the datasets as well as our codes here\footnote{https://anonymous.4open.science/r/ThemeSpecificKG} and will make it public upon acceptance. 
The datasets can be used as a benchmark for theme-specific knowledge graph construction or sub-tasks like entity recognition and open relation extraction.

% \begin{table*}
%     \centering
%     \begin{tabular}{c|c c c|ccc}
%         \hline 
%          & \multicolumn{3}{|c|}{EVB} & \multicolumn{3}{|c}{HAI}\\

%         Method & Entity Recall & Entity Precision & Entity F1 & Entity Recall & Entity Precision & Entity F1  \\
%         \hline
%         OpenIE & 0.62 & 0.36 & 0.46 & 0.52 & 0.28 &0.36\\
%         REBEL & 0.22 & 0.80 &0.35 & 0.16 & 0.87 &0.27\\
%         IMoJIE & 0.44 & 0.49 & 0.46 & 0.33& 0.39 & 0.36\\
%         KG-GPT & 0.72 & 0.69 &0.70 & 0.84 & 0.79 &0.81\\
%         GPT-4 & 0.68 & 0.71 &0.69 & 0.82 & 0.80 &0.83\\
%         \hline
%         TKGCon & \textbf{0.92} & 0.80 & \textbf{0.86} & \textbf{0.90} & \textbf{0.88} &\textbf{0.89}\\
%         \hline
%     \end{tabular}
%     \caption{Comparison with baselines on entity evaluation of KG.}
%     \label{tab:baseline2} 
% \vspace*{-2ex}
% \end{table*}

% \vspace*{-2ex}
\subsection{Evaluation Metrics}
Given the target theme-specific KG, $\mathcal{G}=(\mathcal{E},\mathcal{R},\mathcal{T})$ and the generated KG, $\hat{\mathcal{G}}=(\hat{\mathcal{E}},\hat{\mathcal{R}},\hat{\mathcal{T}})$ with $\mathcal{E} ,\hat{\mathcal{E}}$ denoting the set of gold and recognized entities, $\mathcal{R}, \hat{\mathcal{R}}$ denoting the set of gold and generated relations, and $\mathcal{T}, \hat{\mathcal{T}}$ denoting the set of gold and extracted triples $\left(e_i,r_k,e_j \right)$. We define the following evaluation metrics from different perspectives to evaluate the quality of generated KG.
\begin{itemize}
    \item {\bf Entity Metrics.} We use precision, recall, and F1-score to evaluate the extracted entities. Since KG construction is an open task, there is no penalty on precision if the recognized entities are reasonable but not included in the gold set $\mathcal{E}$. Also, we regard two entities as the same if they are semantically-matched coreference of entities such as \textit{EV battery} and \textit{electrical vehicle batteries}.
    \item {\bf Triple Metrics.} We use the precision, recall, and F1-scores for evaluating the generated triples. Since the relations can be paraphrased into various expressions such as \textit{subclass of} and \textit{type of}, we apply soft matching \cite{jiao2022openvocabulary} by obtaining embeddings of the pre-trained transformer and calculating cosine similarities between the extracted triples and gold triples.
    
    \item {\bf Theme Coherence.} The raw documents usually include noisy information. We expect the KG to focus on theme-related entities and relations. Therefore, we propose a metric to evaluate the theme coherence of triples. We calculate the cosine similarity between triples and the theme descriptions by the pre-trained transformer. The triples with high similarities are seen as theme-coherent triples, thus the metric of theme coherence is defined by the rate of theme-coherent triples in all generated triples.
    %\item {\bf Manually AB Test.[WIP]} Although the similarity-based automatic evaluations can be used as the a reference for semantic correctness, there are still some limitations on identifying which expressions of triples can be better. For example, the automatic evaluations above can hardly identify the better one in \textit{(deep cycle battery, run, forklift)} and \textit{(deep cycle battery, be power source of, forklift)} because they share the same meaning. Therefore, to further evaluate the performance, we compare the extraction results  of different methods and choose the best expression manually based on the following criteria.(1) Salience. A salient relation should have a number of instances in the real world.(2) Ambiguity. A good relation should be accurate in semantics without ambiguity. (3) Informativeness. The relations should be specific and informative based on the context.
\end{itemize}

\vspace*{-2ex}
\subsection{Baselines}
We consider existing open knowledge graph construction methods including two GPT-4 based baselines.
\begin{enumerate}
\item {\bf OpenIE} \cite{10.5555/3061053.3061220} is an open information extraction system which extracts relation triples from plain text. This baseline focuses mainly on the syntactic patterns of the text to compensate for the lack of supervision signals.
\item {\bf REBEL} \cite{huguet-cabot-navigli-2021-rebel-relation} is a generative framework that formulates KG construction as a sequence-to-sequence task that extracts all the triples within the input texts based on BART \cite{lewis2020bart}. The model is pre-trained on the general corpus and evaluated on our corpus.
%\item DocOIE is a document-level open-vocabulary domain-specific knowledge graph construction method.
\item {\bf IMoJIE} \cite{kolluru2020imojie} is a BERT-based OpenIE system that generates information extraction using an iterative Seq2Seq model. We use the model that is trained on the CaRB \cite{bhardwaj2019carb} data.
\item {\bf KnowledgeGraph GPT (KG-GPT)}\footnote{https://github.com/iAmmarTahir/KnowledgeGraphGPT} \cite{kggpt} is a tool that converts plain text to knowledge graph with GPT-4. The approach designs prompts to generate the results in the style of RDF Tuples\footnote{https://www.w3.org/TR/WD-rdf-syntax-971002/}.
\item We also directly take {\bf GPT-4}\footnote{https://openai.com/gpt-4} \cite{openai2021gpt4} as a baseline by prompting it in an end-to-end manner. We use in-context learning to query the theme document one at a time and then combine the generated triples together. The prompt we used is:
\begin{quote}
{\em Given the theme [Theme], extract all knowledge triples in the format of (entity1, relation, entity2) from the following text: [document].}
\end{quote}
\nop{
\vspace*{.5ex}
{\em Extract all  knowledge triples in the format of (entity1, relation, entity2) from the following text: [document].}
\vspace*{.5ex}
}
%\item Grapher \cite{melnyk2022knowledge} is an end-to-end multi-stage KG construction system.
\end{enumerate}
% \begin{itemize}
%     \item 
% \end{itemize}

\begin{figure*}[!ht]
    \centering
    \includegraphics[width=0.98\linewidth]{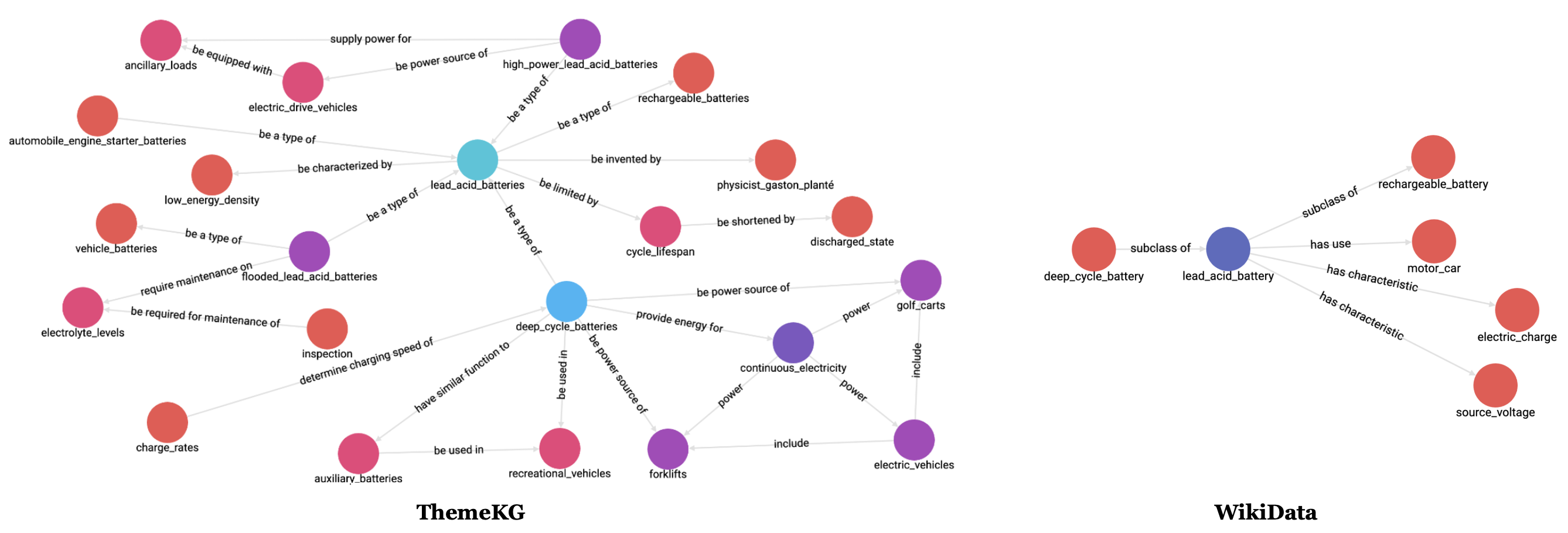}
    \caption{Comparison of \tkg and WikiData on EV battery. {\normalfont For the theme {\em EV battery}, the left side is part of our extracted \tkg extracted from theme-specific document corpus. The triples on the right side are retrieved from WikiData of the same topics. Our \tkg contains more specific entities and relations of the theme compared to the WikiData KG.}}
    \label{fig:case_study1}
\end{figure*}

% \vspace*{-2ex}
\subsection{Main Results Analysis}
We compare our proposed framework \method against baselines on both datasets EVB and HAI. The evaluation results of entities, triples, and theme are present in Table \ref{tab:baseline1}.  

We first evaluate and compare the extracted {\bf entities} in the output KGs of all models. The results in Table \ref{tab:baseline1} show that our \method can outperform other baseline methods considerably on all metrics. Linguistic-based OpenIE has relatively high recall among non-GPT-based baselines. However, the model lacks an understanding of semantics, and thus the entities extracted in OpenIE are usually incomplete, erroneous and noisy, especially in processing long and complex sentences, which leads to low precision and F1-score. End-to-end KGC methods such as KG-GPT and GPT-4 include overly specific entities (such as ``two main types'') and are inflexible to optimize. Compared to end-to-end methods, our \method can greatly improve the recall rate of entities by explicit phrase mining and boost the precision with entity filtering in intermediate steps. 

Our \method also shows decent improvements over the baselines on {\bf triple} extraction and {\bf theme} coherence. Specifically, OpenIE generates relatively poor quality outputs due to its lack of semantic understanding. While supervised sequence generative methods like REBEL and IMoJIE have slightly better precision and theme coherence, they are found weak in recall rate due to the difficulty of generalizing to unseen relations beyond the training dataset, while our datasets include a number of fine-grained and theme-specific relations. GPT-4 is capable of comprehension and summarizing, and thus its induced models (GPT-4 and KG-GPT) have significant improvements compared to others. However, in the setting of \tkg, these two GPT-4-based models may still generate ambiguous and illogical relations or entities (details and examples can be found in Sec. ~\ref{sec:case_study}). Breaking the \tkg construction into multiple stages, our framework \method can effectively improve the recall rate of document-level triples, compared to end-to-end models. With the guidance of theme ontology, \method is able to extract higher quality relations and outperform GPT-based methods in triple F1-score while getting competitive theme coherence on par with GPT-4.

\begin{table*}[htp] \small
\caption{Examples of triples. The bold refers to correct triples.}
\vspace{-2mm}
    \centering
    \begin{tabular}{c|p{15cm}}
       \toprule
       {\bf Method} & {\bf Extracted Triples (on same plain text)} \\
        \midrule
        OpenIE
 & (lead acid batteries, have compared, rechargeable batteries), (It, is, first type), (calendar, impede, lifecycle), (lead acid batteries, have, energy density), (features, make along, their cost), (overall lifespan, is in, discharged state), \textbf{(lead acid batteries, suffer from, cycle lifespan)}, (deep cycle batteries, are, used), (percentage, provide, charge rates)
\\
        \midrule
        GPT-4 & 
\textbf{(lead–acid battery, is a type of, rechargeable batteries)}, 
\textbf{(lead–acid batteries, have, low energy density)},
(lead-acid batteries, have, two main types), 
\textbf{(lead-acid batteries, suffer from, relatively short cycle lifespan)},
(lead-acid batteries, have due to, "double sulfation" in the discharged state),
\textbf{(lead-acid batteries, types, deep cycle batteries)},
(high charge rates, to start, engine), 
\textbf{(flooded lead-acid batteries, are, vehicle batteries)},
\textbf{(deep cycle batteries, require, multi-stage charging)},
(deep cycle batteries, used in, electric vehicles like forklifts or golf carts)
\\
        \midrule
        \method & 
\textbf{(lead acid batteries, be a type of, rechargeable batteries)},
\textbf{(lead acid batteries, be characterized by, low energy density)},
\textbf{(lead acid batteries, be limited by, cycle lifespan)},
\textbf{(cycle lifespan, be shortened by, discharged state)},
\textbf{(flooded lead acid batteries, be a type of, vehicle batteries)},
(automobile engine starter batteries, be different from, deep cycle batteries),
\textbf{(deep cycle batteries, be power source of, forklifts)},
\textbf{(deep cycle batteries, be power source of, golf carts)},
(deep cycle batteries, be used to store, continuous electricity)
 \\
        \bottomrule
    \end{tabular}
    \label{tab:case study}
\end{table*}

\subsection{Ablation Study} \label{sec:ablation}
Essentially, our framework could be separated into two sequential stages: {\em entity recognition (step 1 in Figure~\ref{fig:overview})} and {\em ontology-guided relation extraction (step 2 \& 3 \& 4 in Figure~\ref{fig:overview})}. To validate the effectiveness of the relation ontology, we build a new model variant, named {\em \method (w/o ontology)}, which leverages entity recognition (step 1) from our framework and then directly queries GPT-4 to generate the relations between entities based on the context.
The prompt for querying the relation is as follows:
\begin{quote}
    {\em Extract the relation from [entity1] to [entity2] in the following passage: [context]. Please output in the format of ([entity1], [relation], [entity2]). If no relation from [entity1] to [entity2] is identified based on the context, then output none.} 
\end{quote}
\nop{
\vspace{0.5ex}
{\em Extract the relation from [entity1] to [entity2] in the following passage: [context]. Please output in the format of [entity1], [relation], [entity2]. If no relation is identified, then output none.} 
\vspace{0.5ex}
 }

The performance of \method (w/o ontology) has been added to Table \ref{tab:baseline1}. Note that, our model and this variant use the same entity recognition step, so their entity metrics are all the same. For all triple metrics and theme coherence, this variant model consistently gives lower performance. Without the guidance and restrictions of theme ontology, the variant model generates incorrect relations when directly querying GPT-4. For example, its generated triples \textit{(auxiliary batteries, be used as, recreational vehicles)} and 
\textit{(inspection, requires inspection of, electrolyte levels)} are obviously problematic because the batteries cannot \textit{"be used as"} vehicles and inspection should not have the relation \textit{"requires inspection of"}. 

Also, \method(w/o ontology) tends to give pointless phrases as relations, such as \textit{(charge rates, provide high, golf carts)} and \textit{(electric vehicles, notable exception, Detroit Electric)}. On the contrary, the ontology-based \method can avoid such scenarios with the help of entity ontology and relation ontology, e.g. \textit{(recreational vehicles, be equipped with, auxiliary batteries), (inspection, be required for maintenance of, electrolyte levels)}.

\subsection{Case Study}
\label{sec:case_study}
This section uses the theme \textit{EV battery} as an example and intuitively shows the differences between \tkg and Wikidata. Also, we compare the extracted triples of \method to the results of baselines.

\subsubsection{Compared to Wikidata} Figure ~\ref{fig:case_study1} presents the difference of \tkg and  Wikidata for {\em EV battery}. Due to space limitations, we only show the parts of the triples related to \textit{lead acid batteries} in our constructed \tkg (left side on Figure~\ref{fig:case_study1}). We also retrieve the triples of relevant topics on Wikidata\footnote{See https://www.wikidata.org/wiki/Wikidata:Main\_Page} (right side on Figure~\ref{fig:case_study1}). 

We observe that our constructed \tkg (from raw {\em EV battery} documents) includes more {\bf fined-grained entities} such as \textit{flooded lead acid batteries, automobile engine starter batteries} which are not presented on the right side. 
The \tkg also contains broader and more {\bf fine-grained knowledge triples} (e.g. \textit{lead acid batteries, be limited by, cycle lifespan}) about the theme while Wikidata only includes a few coarse-grained triples. Therefore, we claim that \tkg could provide broader and more in-depth details on specific themes compared to existing KGs.

\subsubsection{Compared to baselines} We also sample some extracted triples from \method and compare them with the outputs from OpenIE, GPT-4 in Table \ref{tab:case study}. 
We could observe that OpenIE misses many entities and triples relevant to the theme since the method is based on linguistic analysis without understanding of the theme. It also leads to pointless triples when the sentences are long and complex. GPT-4 works relatively better due to the strong ability of text understanding. However, directly prompting GPT-4 makes the entities excessively specific (e.g., \textit{"electric vehicles like forklifts or golf carts"}) or leads to ambiguous (e.g., \textit{"is", "have", "were"}) and wrong relations (e.g., \textit{"have due to", "to start"}).

Our \method breaks the \tkg construction task into several steps, which effectively optimizes each step and improves the quality and recall rate of entities. In the example, \method extracts more accurate, e.g., entities \textit{"forklifts"} and \textit{"golf carts"} (compared to the excessively specific entity, \textit{"electric vehicles like forklifts or golf carts"}, generated by GPT-4). Additionally, \method retrieves candidate relations from relation ontology, which prevents relation ambiguity and improves the accuracy in relation extraction. For example, our method generates \textit{"be power source of"} instead of \textit{"used in"} (generated by GPT-4). Furthermore, benefiting from the ontology, \method improves the consistency of relations while relations generated from GPT-4 are oftentimes unstable (e.g. \textit{"is a type of", "types", "are"} for the same kind of relation).

\subsection{Applications of \tkg}
The \tkg are compressed and structured collections of fine-grained knowledge for specific themes. Besides intuitive access to knowledge, \tkg can be used as an external tool to facilitate the downstream tasks. For example, the LLMs may cause hallucinations on the fine-grained details and do not possess up-to-date information for a specific theme. In this case, our \tkg can be retrieved and loaded as tools to augment the LLMs. 

This section tests the model's question-answering (QA) ability over a specific theme. We consider the following models.
\begin{itemize}
    \item \textbf{\tkg-augmented GPT-4 (TKG+GPT4)} uses \tkg (formatted as a list of triples) in GPT-4 for in-context QA since the \tkg is more compressed than the entire theme corpus.
    \item \textbf{Vanilla GPT-4} that directly uses GPT-4 for QA.
    \item \textbf{Retrieval-augmented generation with GPT-4 (RAG+GPT4)} \cite{lewis2020retrieval} is a representative method that retrieving facts from an external knowledge base to augment GPT-4 on the accurate, up-to-date information. We take the raw theme corpus as the external database of RAG for QA. 
\end{itemize}

Table~\ref{tab:application} lists an example of the theme {\em Hamas-Attack-on-Israel (2023)}
(redundant sentences of QA output are all removed). The question is {\em ``Which countries support Hamas or condemn Israel in the Hamas attack on Israel in Oct 2023?''}. To answer the question requires information integration from multiple documents. Apparently, GPT-4 alone cannot deal with the up-to-date information. The RAG+GPT4 retrieves parts of relevant facts while misses information such as text mentioning \textit{Iran} and \textit{Persian Gulf Countries}. Our TKG+GPT4 can directly integrate information from triples, e.g. \textit{(Iran, provide assistance to, Hamas)} and \textit{(North Korea, condemn, Israel)}, without summarizing and reasoning over the documents. Essentially, the \tkg could empower LLMs with timely, fine-grained information and also improve their abilities of reasoning and summarizing over the theme. 

In future work, we will further explore other utilities of ThemeKGs in more scenarios, such as personalized recommendation and tool learning with different ThemeKGs as plugins. 

\begin{table}[tp!] \small
    \centering
    \vspace{-1mm}
\caption{An example of \tkg-augmented GPT-4 on QA task. The bold phrases are the correct answers.}
\vspace{-1.5mm}
    \begin{tabular}{c|p{6cm}}
    \toprule
        {\bf Question}
 & Which countries support Hamas or condemn Israel in the Hamas attack on Israel in Oct 2023?
\\
        \midrule
        Vanilla GPT4 & I'm sorry, but as of my knowledge cutoff date in march 2023, i do not have information on specific events that occurred in october 2023.
\\
        \midrule
        RAG+GPT4 & In the \textbf{Middle East} and \textbf{North Africa}, most countries either condemned Israel or offered full-throated support to Hamas. \textbf{North Korea} is also mentioned as condemning Israel.
\\
        \midrule
        TKG+GPT4 & 1. \textbf{Iran}, 2. \textbf{Persian Gulf countries}, 3. \textbf{North Korea}, 4. most \textbf{Middle East countries}, 5. most \textbf{North Africa countries}
\\
        \bottomrule
    \end{tabular}
    \label{tab:application}
    \vspace{-3.5mm}
\end{table}

%% file: 7-limitation-conclution.tex
\section{Conclusion}
This paper proposes an automated KG construction method (named \method) for constructing a theme-specific knowledge graph (\tkg) from a raw theme corpus. Existing general or domain-specific KGs could be limited in information granularity and timeliness, while our focused \tkg contains fine-granular, timely, theme-specific entities and relations, providing more theme-relevant details (such as specialized terms or key events in recent news) for downstream applications. Our approach leverages well-curated theme ontology information in Wikipedia as well as the strong reasoning ability of LLM for relation candidate generation. Experiments on {\em EV battery} and {\em Hamas-attack-on-Israel} show that \method can automatically generate fine-grained, high-quality \tkg in comparison with various existing KG construction approaches.